\documentclass[11pt,a4paper]{article}
\usepackage{authblk}
\usepackage{emnlp2017new}
\usepackage{times}
\usepackage{latexsym}
\usepackage{amsmath}
\usepackage{graphicx}
\usepackage{amsfonts}
\usepackage[utf8]{inputenc}

\emnlpfinalcopy

\title{Past, Present, Future: A Computational Investigation of the
  Typology of Tense in 1000 Languages}

\date{}

\newcounter{notecounter}
\newcommand{\enotesoff}{\long\gdef\enote##1##2{}}
\newcommand{\enoteson}{\long\gdef\enote##1##2{{
\stepcounter{notecounter}
{\large\bf
\hspace{1cm}\arabic{notecounter} $<<<$ ##1: ##2
$>>>$\hspace{1cm}}}}}
\enoteson
\enotesoff

\def\dnrm#1{\mbox{$_{\hbox{\scriptsize #1}}$}}

\def\figref#1{Figure~\ref{fig:#1}}
\def\figlabel#1{\label{fig:#1}\label{p:#1}}

\def\tabref#1{Table~\ref{tab:#1}}
\def\tablabel#1{\label{tab:#1}\label{p:#1}}

\def\secref#1{\S\ref{sec:#1}}
\def\seclabel#1{\label{sec:#1}\label{p:#1}}
\def\eqref#1{Eq.~\ref{eqn:#1}}

\author[1,2]{Ehsaneddin Asgari}
\author[1]{Hinrich Sch\"{u}tze}
\affil[1]{Center for Information and Language Processing, LMU Munich, Germany}
\affil[2]{Applied Science and Technology, University of California, Berkeley, CA, USA\\\protect\\ {inquiries@cislmu.org}}

\begin{document}

\maketitle

\begin{abstract}
We present SuperPivot, an analysis method for low-resource languages
that occur in a superparallel corpus, i.e., in a corpus that
contains an order of magnitude more languages than parallel
corpora currently in use.
We show that SuperPivot
performs well 
  for the crosslingual analysis of the linguistic phenomenon
  of tense. We produce analysis results for more than 1000
  languages, conducting -- to the best of our knowledge -- the largest crosslingual computational study
  performed to date.
We extend existing methodology for leveraging parallel
corpora for typological analysis by overcoming a limiting
assumption of earlier work:
We only require that a linguistic feature is overtly marked
in a \emph{few} of thousands of languages as opposed to requiring
that it be marked in \emph{all} languages under investigation.
\end{abstract}

\section{Introduction}
Significant linguistic resources such as machine-readable
lexicons and part-of-speech (POS) taggers are available for at
most a few hundred languages. This means that the majority
of the languages of the world are low-resource.
Low-resource languages like Fulani are spoken by tens of
millions of people and are politically and economically
important; e.g., to manage a sudden refugee crisis, NLP
tools would be of great benefit. Even ``small'' languages
are important for the preservation of the common heritage of
humankind that includes natural remedies and linguistic and
cultural diversity that can potentially enrich everybody.
Thus, developing analysis methods for low-resource languages
is one of the most important challenges of NLP today.

We address this challenge by proposing a new method for
analyzing what we call \emph{superparallel corpora}, corpora
that are by an order of magnitude more parallel than corpora
that have been available in NLP to date. The corpus we work
with in this paper is the Parallel Bible Corpus (PBC) that
consists of translations of the New Testament in 1169
languages. Given that no NLP analysis tools are available
for most of these 1169 languages, how can we extract the
rich information that is potentially hidden in such
superparallel corpora?

The method we propose is based on two hypotheses.
\textbf{H1 Existence of overt encoding.} For any important linguistic distinction $f$ that is
  frequently encoded across languages in the world, there
  are a few languages that encode $f$ overtly on the surface.
 \textbf{H2 Overt-to-overt and overt-to-non-overt
  projection.} For a language $l$ that encodes $f$, 
a projection of $f$ from the
  ``overt languages'' to $l$ in the superparallel corpus
  will identify the  encoding that $l$ uses for $f$, both in
  cases in which the encoding that $l$ uses is overt and in
  cases in which the encoding that $l$ uses is non-overt.
Based on these two hypotheses, our method proceeds in 5 steps.

\textbf{1. Selection of a linguistic feature.}
We select  a linguistic feature $f$ of interest. 
Running example: We select past
tense as feature $f$.

\textbf{2. Heuristic search for head pivot.}
Through a heuristic search, we find 
a language $l^h$ that contains
a \emph{head pivot} $p^h$ that is highly correlated with the
linguistic feature of interest. 

Running example: "ti" in
Seychelles Creole (CRS).
CRS ``ti'' meets our requirements for a head pivot well as will  be verified empirically in \secref{exp}.
First, "ti" is a surface marker: it is easily
identifable through whitespace tokenization and it is not
ambiguous, e.g., it does not have a second meaning apart
from being a grammatical marker. Second, "ti" is
a good marker for  past tense in terms of both
``precision'' and ``recall''. CRS has mandatory
past tense marking (as opposed to languages in which 
tense marking is facultative) and "ti" is highly correlated
with the general notion of past tense. 

This does not
mean that every clause that a linguist would regard as past
tense is marked with "ti" in CRS. For example, some tense-aspect
configurations that are similar to English present perfect
are marked with "in" in CRS, not with "ti" (e.g., 
ENG ``has commanded'' is translated as ``in ordonn'').

Our goal is not to find a head
language and a head pivot
that is a perfect marker of $f$. Such a head pivot probably
does not exist; or, more precisely, linguistic features are
not completely rigorously defined. In a sense, one of the
significant contributions of this work is that we provide
more rigorous definitions of past tense across languages;
e.g., "ti" in CRS is one such rigorous definition of past
tense and it automatically extends (through projection) to
1000 languages in
the superparallel corpus.

\textbf{3. Projection of head pivot to larger pivot set.} Based on
an alignment of the head language to the other languages in
the superparallel corpus, we project the head
pivot to all other languages and search for highly
correlated surface markers, i.e., we search for additional
pivots in other languages. This projection to more pivots
achieves three goals. First, it makes the method more
\emph{robust}. Relying on a single pivot would result in many
errors due to the inherent noisiness of linguistic data and
because several components we use (e.g., alignment of the languages
in the superparallel corpus) are imperfect. Second, as we
discussed above, the head pivot does not necessarily have
high ``recall''; our example was that CRS ``ti'' is not
applied to certain clauses that would be translated using
present perfect in English. Thus, moving to a larger pivot
set \emph{increases recall}. Third, as we will see below, the pivot
set can be leveraged to create a \emph{fine-grained map of the
linguistic feature}. Consider clauses referring
to eventualities in the past that English speakers would
render in past progressive, present perfect and simple past
tense. Our hope is that the pivot set will cover these
distinctions, i.e., one of the pivots marks past
progressive, but not present prefect and simple past,
another pivot marks present perfect, but not the other two
and so on. It is beyond the scope of this paper to
verify that we can produce such an analysis for all
linguistic features,
but a promising example of this type of map, including
distinctions like progressive and perfective aspect,
is given
in \secref{map}.

Running example: We compute the correlation of ``ti'' with
words in other languages and select the 100 highest
correlated words as pivots. Examples of pivots we find this
way are Torres Strait Creole ``bin'' (from English ``been'')
and Tzotzil ``laj''. ``laj'' is a perfective marker, e.g.,
``Laj meltzaj -uk'' `LAJ be-made subj' means ``It's done
being built'' \cite{aissen1987tzotzil}.

\textbf{4. Projection of  pivot set to all languages.} Now that
we have a large pivot set, we project the pivots to all
other languages to search for linguistic devices that
express the linguistic feature $f$. Up to this point, we
have made the assumption that it is easy to segment
text in all languages into pieces of a size that is not too
small (individual characters of the Latin alphabet would be
too small) and not too large (entire sentences as tokens
would be too large). Segmentation on
standard delimiters is a good approximation for the majority
of languages -- but not for all: it undersegments some
(e.g., the polysynthetic language Inuit) and oversegments others
(e.g.,  languages that use punctuation
marks as regular characters). 

For this reason, we do not
employ tokenization in this step. Rather we search for
character $n$-grams ($2 \leq n \leq 6$) to find linguistic
devices that express $f$. This implementation of the search
procedure is a  limitation -- there are many linguistic
devices that cannot be found using it, e.g., templates in
templatic morphology. We 
leave addressing this for future work
(\secref{future}).

Running example: We find ``-ed'' for English
and ``-te'' for German as surface features that are highly
correlated with the 100 past tense pivots.

\textbf{5. Linguistic analysis.} 
The result of the previous steps is a superparallel corpus
that is richly annotated with information about linguistic
feature $f$. This structure can be exploited for \emph{the
analysis of a single language} $l^i$ that may be the focus of
a linguistic investigation. Starting with the character n-grams
that were found
in the step ``projection of  pivot set to all languages'',
we can explore their use and function, e.g, for the
mined n-gram ``\mbox{-ed}'' in English (assuming English is the language $l^i$ and  it is unfamiliar to us). Many of the
other 1000 languages provide annotations of linguistic
feature $f$ for $l^i$: both the languages that are part of
the pivot set (e.g., Tzotzil ``laj'') and  the mined n-grams in other
languages that we may have some knowledge of (e.g., ``-te''
in German).

We can also use the structure we have generated for
\emph{typological analysis across languages} following the work of
Michael Cysouw. He has pioneered a new methodology for
typology (\cite{cysouw2014inducing},
\secref{related}). We do not contribute any innovations to
typology in this paper, but our method is a significant
advancement computationally over Cysouw's  work
because we overcome many of his limiting
assumptions. Most importantly, our
method scales to thousands of languages as we demonstrate
below whereas Cysouw worked on a few dozen.

Running example: We sketch  the type of analysis
that our new method makes possible in \secref{map}.

The above steps 
``1. heuristic search for head pivot'' and
``2. projection of head pivot to larger pivot set'' are
based on H1: we assume the \textbf{existence of overt
coding} in a subset of languages.

The above steps
``2. projection of head pivot to larger pivot set'' and
``3. projection of  pivot set to all languages'' are based on
H2: we assume that \textbf{overt-to-overt and overt-to-non-overt projection}
is possible.

In the rest of the paper, we will refer to the method that consists of steps 1 to 5 as
\emph{SuperPivot}:
``linguistic analysis of SUPERparallel corpora using surface PIVOTs''.

We make three contributions.
(i) Our basic hypotheses are H1 and H2.
(H1) For an important linguistic feature, there exist a few languages
that mark it overtly and easily recognizably. (H2) It is
possible to project overt markers to overt and non-overt
markers in other languages. Based on these two hypotheses 
we design SuperPivot, a new method for analyzing
  highly parallel corpora, and show that it performs well
  for the crosslingual analysis of the linguistic phenomenon
  of tense.
(ii) Given a superparallel corpus, SuperPivot can be used
  for the analysis of \emph{any low-resource language}
  represented in that corpus. In the
  supplementary material, we present results of our analysis
  for three tenses (past, present, future) for
  1163\footnote{We exclude six of the 1169 languages because
    they do not share enough verses with the rest.}
  languages. An evaluation of accuracy is presented
  in \tabref{mrr}.
(iii) We extend Michael Cysouw's pioneering work on
  typological analysis using parallel corpora by overcoming
  several limiting factors. The most important is that
  Cysouw's method is only applicable if markers of the
  relevant linguistic feature are recognizable on the
  surface in \emph{all} languages. In contrast, we only
  assume that 
 markers of the
  relevant linguistic feature are recognizable on the
  surface in \emph{a small number of} languages.

\section{SuperPivot: Description of method}
\seclabel{superpivot}
\textbf{1. Selection of a linguistic feature.} The
linguistic feature of interest $f$ is selected by 
the person 
who performs a SuperPivot
analysis, i.e., by a linguist, NLP researcher or data
scientist. Henceforth, we will refer to this person as the linguist.

In this paper, $f \in F = \{ \mbox{past}, \mbox{present}, \mbox{future}\}$.

\textbf{2. Heuristic search for head pivot.}  There are
several ways for finding the head language and the head
pivot. Perhaps the linguist knows a language that has a good
head pivot. Or she is a trained typologist and can find the
head pivot by consulting the typological literature.

In this paper, we use our knowledge of English and an
alignment from English to all other languages to find 
head pivots. (See below for details on alignment.)
We define a ``query'' in English and search
for words that are highly correlated to the query in other
languages. For future tense, the query is simply the word
``will'', so we search for words in other languages that are
highly correlated with ``will''. For present tense, the
query is the union of ``is'', ``are'' and ``am''. So we
search for words in other languages that are highly
correlated with the ``merger'' of these three words. For
past tense, we POS tag the English part of PBC
and merge all words tagged as past tense into one past tense
word.\footnote{Past tense is defined as  tags
BED, BED*, BEDZ, BEDZ*,
  DOD*, VBD, DOD. We use NLTK
\cite{bird2006nltk}.}
We then search for words in other languages that are
highly correlated with this artificial past tense word.

As an additional constraint, we do not select
the most highly correlated word as the head pivot, but the
most highly correlated word in a Creole language. Our
rationale is that Creole languages are more regular
than other languages because they are young and have not
accumulated ``historical baggage'' that may make computational
analysis more difficult.

\tabref{topmark} lists the
three head pivots for $F$.

\textbf{3. Projection of head pivot to larger pivot set.}
We first use fast\_align \cite{dyer2013simple} to align the
head language to all other languages in the corpus.  This
alignment is on the word level.  

We compute a score
for each word in each language based on the number of times
it is aligned to the head pivot, the number of times it
is aligned to another word and the total frequencies of head
pivot and word. We use $\chi^2$
\cite{casella08statistical} as the score throughout this paper.
Finally, we select the $k$ words as pivots that have the
highest association score with the head pivot. 

We impose the
constraint that we only select one pivot per language. So as
we go down the list, we skip pivots from languages for which
we already have found a pivot. We set $k=100$ in this
paper. \tabref{topmark}
gives the top 10 pivots.

\seclabel{projection2all}
\textbf{4. Projection of  pivot set to all languages.} As
discussed above,
the process so far has been
based on tokenization. To be able to find markers that 
cannot be easily detected on the surface (like ``-ed'' in
English), we identify non-tokenization-based character n-gram features in
step 4.

The immediate challenge is that without tokens, we have no
alignment between the languages anymore. 
We could simply assume that the occurrence of a pivot has
scope over the entire verse. But this is clearly inadequate,
e.g.,  for the sentence 
 ``I arrived yesterday,
I’m staying today, and I will leave tomorrow'', it is
incorrect to say that it is marked as past tense 
(or future tense)
in its
entirety.
Fortunately, the
verses in the New Testament mostly have a
simple structure that limits the variation in where a
particular piece of content occurs in the verse. We
therefore make the assumption that a particular relative
position in language $l_1$ (e.g., the character at relative
position 0.62) is aligned with the same relative position in
$l_2$ (i.e., the character at relative position
0.62). This is likely to work for a simple example like
 ``I arrived yesterday,
I’m staying today, and I will leave tomorrow'' across languages.

In our analysis of errors, we found many cases where
this assumption breaks down. A well-known problematic
phenomenon for our method is the difference between, say,
VSO and SOV languages: the first class puts the verb at the
beginning, the second at the end. However, keep in mind that
we accumulate evidence over $k=100$ pivots and then compute
aggregate statistics over the entire corpus. As our
evaluation below shows, the ``linear alignment'' assumption
does not seem to do much harm given the general robustness
of our method.

One design element that increases robustness is that we find
the two positions in each verse that are most highly
(resp.\ least highly) correlated
with the linguistic feature $f$.
Specifically, we compute the
relative position $x$ of each pivot that occurs in the
verse and  apply a Gaussian filter ($\sigma=6$ where the
unit of length is the character), i.e., we
set $p(x) \approx 0.066 $ (0.066 is the density of a
Gaussian with $\sigma=6$ at $x=0$)
and center a bell curve around $x$. The total score for a
position $x$ is then the sum of the filter values at $x$
summed over all occurring pivots. Finally, we select the
positions $x\dnrm{min}$ and  $x\dnrm{max}$ with lowest and highest
values for each verse.

$\chi^2$ is then computed based on the number of times a
character n-gram occurs in a window of size $w$ around $x\dnrm{max}$
(positive count) 
and 
in a window of size $w$ around $x\dnrm{min}$ (negative count). Verses in
which no pivot occurs are used for the negative count in
their entirety. The top-ranked character n-grams are then
output for analysis by the linguist.
We set $w=20$.

\textbf{5. Linguistic analysis.} 
We now have created a structure that contains rich
information about  the linguistic feature: for each verse we
have relative positions of pivots that can be projected
across languages. We also have maximum positions within a
verse that allow us to pinpoint the most likely place in
the vicinity of which linguistic feature $f$ is marked in
all languages. This
structure can be used for the analysis of individual
low-resource languages as well as for typological
analysis. We will give an example of such an analysis in \secref{map}.

\textbf{6. Hierarchical clusterings of markers and languages.} As an additional evaluation, we worked on hierarchical clusterings of past, present and future pivots. As detailed in  \secref{projection2all}.4, we represent each verse by a vector of length 100 showing which pivot markers are used to express this verse. The other way of looking at these data is that for each marker we have an occurrence distribution over verses and we may exploit these data to demonstrate the distance between markers.  For the purpose of comparing two markers, we propose calculation of the Jensen-Shannon divergence between the normalized occurrence distribution over verses:
\[
D_{m_{p_i},m_{p_j}}=JSD(\hat{m_{p_i}},\hat{m_{p_j}}),\vspace{-0.25cm}
\]
where $\hat{m_{p_i}}$ and $\hat{m_{p_i}}$, are the normalized occurrence distributions over verses. We compare the obtained distance between markers with genetic distance of their corresponding languages using WALS information~\cite{dryer2005world}. For visualization purposes, we perform Unweighted Pair Group Method with Arithmetic Mean (UPGMA) hierarchical clustering on the pairwise distance matrix of the marker for each tense separately \cite{johnson1967hierarchical}. 

In addition to clustering of pivot markers for each tense
separately, we performed the same comparison for all top
markers of 1107 languages\footnote{We exclude languages that
  have fewer than 
  7000  verses in common with the pivot language to ensure
  quality of marker.}  and take the average
distances of languages in past, present, and future
marking. This allows us to compare the average tense
behavior of languages.
\[
D_{l_i,l_j}=\frac{1}{3} (JSD_{past}+JSD_{present}+JSD_{future}),\vspace{-0.25cm}
\]

\section{Data, experiments and results}
\seclabel{exp}
\subsection{Data}
We use a New Testament subset of the Parallel Bible Corpus (PBS)
\cite{mayer2014creating} that
consists of 1556 translations 
of the the Bible
in
1169 unique languages. We consider two languages to be
different if they have different ISO 639-3
codes.

The translations are aligned on the verse
level. However, many
translations do not have complete coverage, so that most
verses are not present in at least one translation.
One reason for this is that
sometimes several consecutive verses are 
merged, so that one verse contains material that is in
reality not part of it and the merged verses may then be
missing from the translation.
Thus, there is a trade-off between
number of parallel translations and number of verses they
have in common. Although 
some preprocessing was done by the authors of the resource,
many translations are not preprocessed. For example,
Japanese is not tokenized.
We also observed some incorrectness and sparseness in
the metadata. One example is that
one Fijian
translation (see \secref{map}) is tagged
fij\_hindi, but it is Fijian, not Fiji Hindi.

We use the 7958 verses with the best coverage across languages.

\subsection{Experiments}
\textbf{1. Selection of a linguistic feature.}
We conduct three experiments for the linguistic features
past tense, present tense and future tense.

\textbf{2. Heuristic search for head pivot.}
We use the queries described in \secref{superpivot} for
finding the following three head pivots. (i)
Past tense head pivot: ``ti'' in Seychellois Creole (CRS)
\cite{mcwhorter2005defining}. (ii)
Present tense head pivot: ``ta'' in Papiamentu (PAP)
\cite{andersen1990papiamentu}.
(iii)
Future tense head pivot: ``bai'' in Tok Pisin (TPI)  \cite{traugott1978expression,sankoff1990grammaticalization}.

\def\markerspace{0.075cm}

\begin{table*}
\centering 
\small
\begin{tabular}{ll@{\hspace{\markerspace}}l@{\hspace{\markerspace}}l|l@{\hspace{\markerspace}}l@{\hspace{\markerspace}}l|l@{\hspace{\markerspace}}l@{\hspace{\markerspace}}l} 
&\multicolumn{3}{c}{past} & \multicolumn{3}{c}{present} & \multicolumn{3}{c}{future} \\ 
&code & language & pivot & code & language & pivot & code &
language & pivot   \\\hline\hline
head pivots & CRS&  Seychelles C.  & $ti$ &PAP& Papiamentu  & $ta$ &TPI& Tok Pisin  & $bai$ \\\hline
&GUX&  Gourmanchéma  & $den$ &NOB& Norwegian Bokmål& $er$ &LID& Nyindrou  & $kameh$ \\
&MAW&  Mampruli  & $daa$ &HIF& Fiji Hindi & $hei$ &GUL& Sea Island C.  & $gwine$ \\
&GFK&  Patpatar  & $ga$ &AFR& Afrikaans  & $is$ &TGP& Tangoa  & $pa$ \\
&YAL&  Yalunka  & $yi$ &DAN& Danish  & $er$ &BUK& Bugawac  & $oc$ \\
&TOH&  Gitonga  & $di$ &SWE& Swedish  & $\ddot{a}r$ &BIS& Bislama  & $bambae$\\
&DGI&   Northern Dagara  & $t\iota$ &EPO& Esperanto  & $estas$ &PIS& Pijin  & $bae$\\
&BUM&  Bulu (Cameroon) & $nga$ &ELL& Greek  & $\epsilon\acute{\iota}\nu\alpha\iota$ &APE& Bukiyip  & $eke$ \\
&TCS&  Torres Strait C.  & $bin$ &HIN& Hindi  & $haai$ &HWC& Hawaiian C.  & $goin$\\
&NDZ&  Ndogo  & $gi\grave{\iota}$   &NAQ& Khoekhoe  & $ra$ &NHR& Nharo  & $gha$ 
\end{tabular}
\caption{Top ten past, present, and future tense pivots
  extracted from 1163 languages. C. = Creole\tablabel{topmark}} 
\end{table*}

\textbf{3. Projection of head pivot to larger pivot set.}
Using  the method described in \secref{superpivot}, we
project each head pivot to a set of $k=100$ pivots.
\tabref{topmark} gives 
the top
10 pivots for each tense.

\textbf{4. Projection of  pivot set to all languages.} 
Using  the method described in \secref{superpivot}, we
compute highly correlated character $n$-gram features, $2
\leq n \leq 6$, for all
1163 languages.

See \secref{map} for the last step of SuperPivot:
\textbf{5. Linguistic analysis.}

\def\mrrspace{0.2cm}

\begin{table}
\small
\centering 
\begin{tabular}{l@{\hspace{\mrrspace}}||l@{\hspace{\mrrspace}}l@{\hspace{\mrrspace}}l@{\hspace{\mrrspace}}|l@{\hspace{\mrrspace}}}
 language  & past & present   & future  & all\\  \hline\hline
 Arabic &  1.00& 0.39& 0.77 & 0.72\\
 Chinese & 0.00& 0.00& 0.87 & 0.29\\
 English & 1.00& 1.00& 1.00 & 1.00\\
 French & 1.00& 1.00& 1.00 & 1.00\\
 German & 1.00& 1.00& 1.00 & 1.00\\
 Italian & 1.00& 1.00& 1.00 & 1.00\\
 Persian & 0.77& 1.00& 1.00 & 0.92\\
 Polish & 1.00& 1.00& 0.58 & 0.86\\
 Russian & 0.90& 0.50& 0.62 & 0.67\\
 Spanish & 1.00& 1.00& 1.00 & 1.00\\\hline
 all & 0.88& 0.79& 0.88 & 0.85
\end{tabular}
\tablabel{mrr} 
\caption{MRR results for step 4. See text for details.}
\end{table}

\subsection{Evaluation}
We rank n-gram features and retain the top 10,
for each linguistic feature, for each
language and for each n-gram size. We process
1556 translations. Thus, in total, we extract $1556
\times 5 \times 10$ n-grams.

\tabref{mrr} shows Mean Reciprocal Rank (MRR) for 10 languages. The rank for a
particular ranking of n-grams  is the first n-gram
that is highly correlated with the relevant tense; e.g.,
character subsequences of the name ``Paulus''
are evaluated as incorrect,
the subsequence ``-ed'' in English as correct for past. 
MRR is averaged over all n-gram sizes,  $2
\leq n \leq 6$.
Chinese has
consistent tense marking only for future, so results are
poor. Russian and Polish perform poorly because their
central grammatical category is aspect, not tense.
The poor performance on Arabic is
due to the limits of character n-gram features for a
``templatic'' language.

During this evaluation, we noticed a surprising
amount of variation within translations of one language;
e.g., top-ranked n-grams for some German translations
include names like ``Paulus''. We suspect that for literal
translations, linear alignment (\secref{superpivot}) 
yields good n-grams. But many translations are
free, e.g., they change the sequence of clauses. This
deteriorates mined n-grams. See \secref{future}.

\textbf{Hierarchical clusterings of markers.} Hierarchical
clusterings of past, present and future pivots using JSD
between the normalized occurrence distribution over verses
are shown in \figref{pt_marker}, \figref{present_marker},
and \figref{future_marker} for past, present, and future
tenses respectively. In addition to markers clusterings, the
average tense behavior clustering of 1107 languages is
depicted in \figref{lang_clust}. In these figures languages
are colored based on their language families using
WALS~\cite{dryer2005world}, languages without family
information on WALS are uncolored. We observed that most of
pivot past and future markers belong to Niger Congo family
and present markers are mostly within Indo-European
family. It can be seen that in many cases languages with
the same family behave accordingly in tense marking. For
instance, in past tense marking Oto-Manguean languages use
almost the same marker of $ni$ with small writing variations
(\figref{pt_marker}). Although Tezoatlán Mixtec did not have
a record on WALS, since its marker is the same as other
Oto-Manguean languages and works almost identical to $ni$ in
Oto-Manguean languages, we may guess this language is also
Oto-Manguean, which turned out to be true when we performed
further
searches.\footnote{https://www.ethnologue.com/language/mxb}
There were many
of such cases for which we could guess the family of
language based on their tense marking similarities in
\figref{pt_marker}, \figref{present_marker} and
\figref{future_marker}.

We use normalized JSD $(0 \leq JSD \leq 1)$ for
comparison of each pair of languages/markers; this allows us to
investigate whether a simple threshold of $0.5$ 
accurately predicts
whether two languages are genetically related or
not. The results are summarized in \tabref{jsd}. Although
the average tense marking divergence has a low recall, it
expresses a high precision of 0.36, where the random chance
is $\frac{1}{103}\approx0.01$. Thus, it means that if
divergence of tense marking is low the languages are very
likely to be genetically related. This conclusion is
supported by
\figref{lang_clust} where  many small clusters of
nodes have the same color. This suggests that our method may
help in completion of WALS.

\def\jsdspace{0.2cm}

\begin{table*}
\small
\centering 
\begin{tabular}{l@{\hspace{\jsdspace}}||l@{\hspace{\jsdspace}}|l@{\hspace{\jsdspace}}|l@{\hspace{\jsdspace}}|l@{\hspace{\jsdspace}}}
   &   avg tense  & past  & present  & future   \\  
   &    ($\frac{696}{1107}$ lang. - 103 fam.) & ($\frac{55}{100}$ lang. - 15 fam.) &  ($\frac{70}{100}$ lang. - 17 fam.) &($\frac{44}{100}$ lang. - 16 fam.)   \\  \hline\hline
 accuracy & 0.93 & 0.55 & 0.81 & 0.58\\
 precision & 0.36 & 0.18 & 0.75 & 0.16\\
 recall & 0.01 & 0.59 & 0.37 & 0.61\\
 TNR & 0.99 & 0.55 & 0.96 & 0.58\\
\end{tabular}
\tablabel{jsd} 
\caption{Language family similarity prediction results based
  on coordinated marking of verses. Only languages with
  records on WALS are included in this evaluation.
TNR: true negative rate.}
\end{table*}

\begin{figure*}[ht]
\begin{tabular}{ll}
 \includegraphics[width=0.9\textwidth,trim={0cm 4.7cm 0cm 2.56cm},clip]{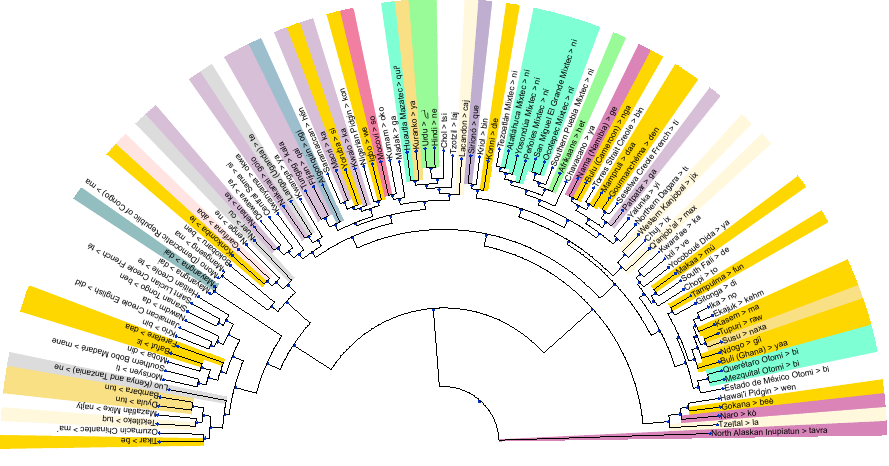}
&
 \includegraphics[width=0.1\textwidth,trim={7.5cm 9cm 0cm 0cm},clip]{Past_markers.pdf}
\end{tabular}
  \caption{\figlabel{pt_marker} Clustering of 100 pivot past tense markers. Each node is colored based on its family information. Languages with no record on WALS remained white. This clustering is based on JSD of markers in marking 5960 verses in bible. We observed that most of
pivot past and future markers belong to Niger Congo family and present markers are mostly within Indo-European
family. It can be seen that in many cases languages with the same family behave accordingly in tense marking. For instance, in past tense marking Oto-Manguean languages use almost the same marker of $ni$ with small writing variations
(\figref{pt_marker}). Although Tezoatlán Mixtec did not have a record on WALS, since its marker is the same as other
Oto-Manguean languages and works almost identical to $ni$ in Oto-Manguean languages, we may guess this language is also Oto-Manguean, which turned out to be true when we performed further searches.}
\end{figure*}

\begin{figure*}[ht]
\begin{tabular}{ll}
 \includegraphics[width=0.9\textwidth,trim={0cm 5.2cm 0cm 2.92cm},clip]{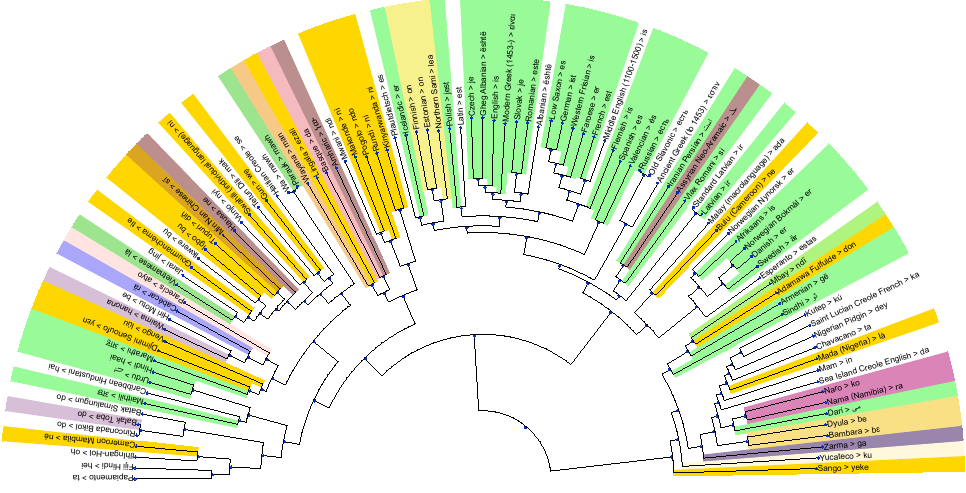}
&
 \includegraphics[width=0.1\textwidth,trim={8.1cm 9.5cm 0cm 0cm},clip]{Present_markers.pdf}
\end{tabular}
  \caption{\figlabel{present_marker} Clustering of 100 pivot present tense markers. Each node is colored based on its family information. Languages with no record on WALS remained white. This clustering is based on JSD of markers in marking 6590 verses in bible. It can be seen that in many cases languages with the same family behave accordingly in tense marking.}
\end{figure*}

\begin{figure*}[ht]
\begin{tabular}{ll}
 \includegraphics[width=0.9\textwidth,trim={0cm 4.7cm 0cm 2.76cm},clip]{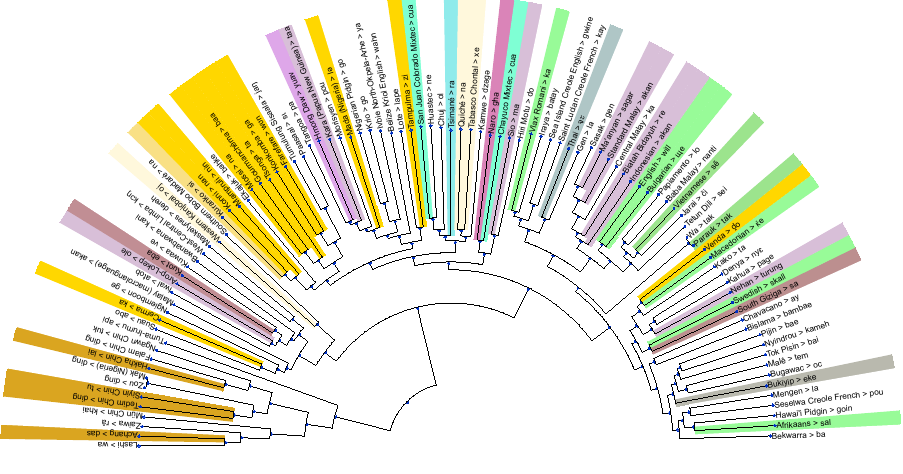}
&
 \includegraphics[width=0.1\textwidth,trim={7.6cm 9cm 0cm 0cm},clip]{Future_markers.pdf}
\end{tabular}
  \caption{\figlabel{future_marker} Clustering of 100 pivot future tense markers. Each node is colored based on its family information. Languages with no record on WALS remained white. This clustering is based on JSD of markers in marking 5733 verses in bible. It can be seen that in many cases languages with the same family behave accordingly in tense marking.}
\end{figure*}

\begin{figure*}[ht]
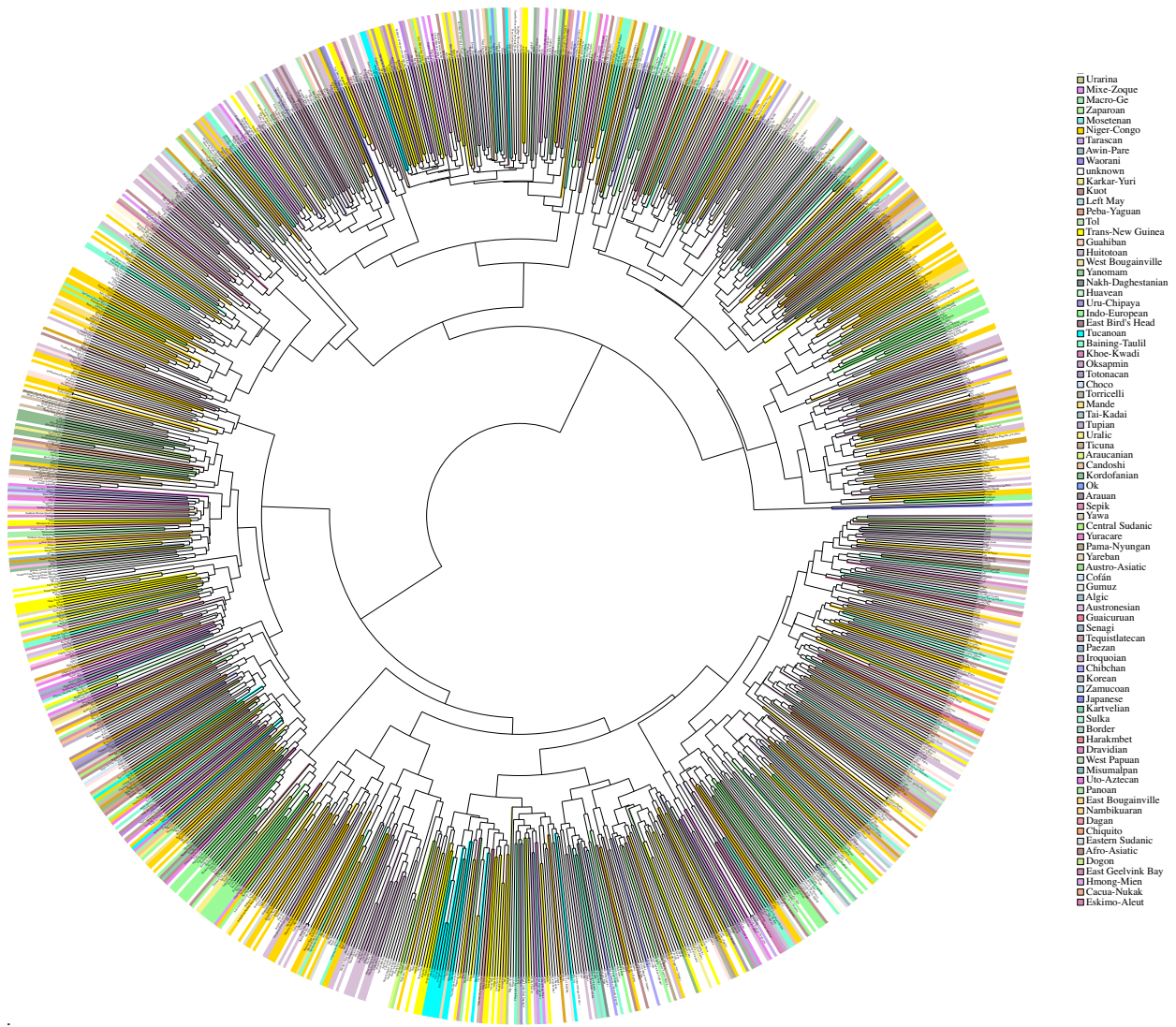

\begin{tabular}{ll}
 \includegraphics[width=0.9\textwidth,trim={0.01cm 0.1cm 0cm 0.01cm},clip]{Tense_circle.pdf}
&
 \includegraphics[width=0.1\textwidth,trim={120cm 120cm 0cm 20cm},clip]{legend.pdf}
\end{tabular}
  \caption{\figlabel{lang_clust} Clustering of 1107 languages based on the average Jensen-Shannon divergence in past, present, and future marking of their respective top markers. Each node is colored based on its family information. Languages with no record on WALS remained white. It can be seen that many small clusters of nodes have the same color, which together with our quantitative evaluation supports that if divergence of tense marking is low the languages are very likely to be genetically related.}
\end{figure*}

\begin{figure*}
\centering
  \includegraphics[width=1\textwidth,trim={11cm 6cm 10.5cm 5.5cm},clip]{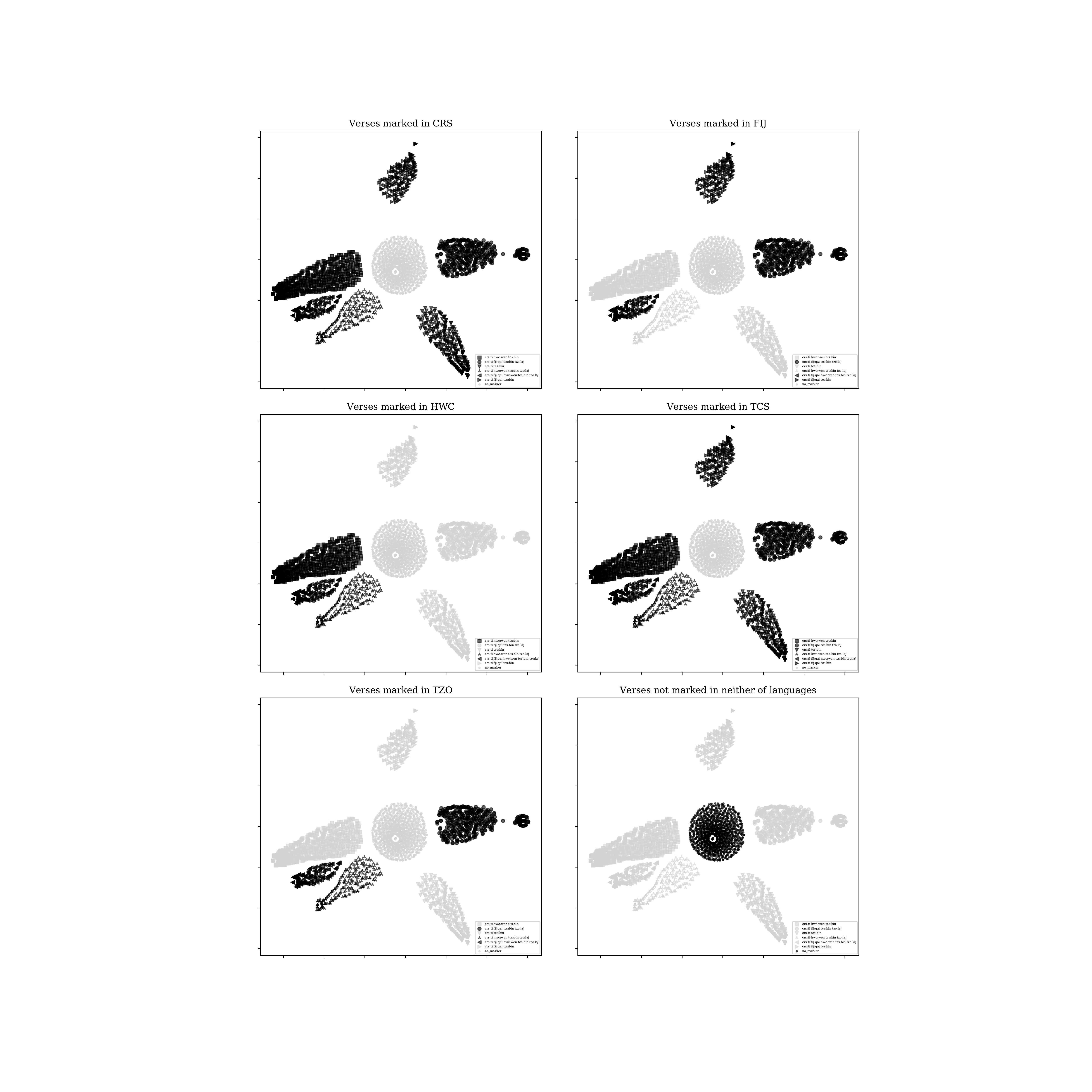}
  \caption{\figlabel{pt} A map of past tense based on the
    largest clusters of verses 
with particular combinations of the past tense pivots from
Seychellois Creole (CRS), Fijian (FIJ), Hawaiian Creole
(HWC), Torres Strait Creole (TCS) and Tzotzil (TZO).
For each of the five languages, we present a subfigure
that highlights the subset of verse clusters that are marked
by the pivot of that language. The sixth subfigure
highlights verses not marked by any of the five pivots.}
\end{figure*}

\section{A map of past tense}
\seclabel{map}
To illustrate the potential of our method we select five out
of the 100 past tense pivots that give rise to large
clusters of distinct combinations. Starting with CRS, we
find other pivots that ``split'' the set of verses that
contain the CRS past tense pivot ``ti'' into two parts that
have about the same size. This gives us two sets. We now
look for a pivot that splits one of these two sets about
evenly and so on. After iterating four times, we arrive at
five pivots: CRS ``ti'', Fijian (FIJ)  ``qai'', Hawaiian
Creole (HWC)
``wen'', Torres Strait Creole (TCS) ``bin'' and Tzotzil (TZO)
``laj''.

\figref{pt} shows a t-SNE 
\cite{maaten2008visualizing}
visualization of the large clusters
of combinations that are found for these five languages,
including one cluster of verses that do not contain any of
the five pivots. 

This figure is a map of past tense for all
1163 languages, not just for CRS, FIJ, HWC, TCS and TZO:
once the interpretation of a particular cluster has been
established based on CRS, FIJ, HWC, TCS and TZO, we can
investigate this cluster in the 1164 other languages by
looking at the verses that are members of this cluster. This
methodology supports the empirical investigation of questions like
``how is progressive past tense expressed in language X''?
We just need to look up the cluster(s) that correspond to
progressive past tense, look up the verses that are members
and retrieve the text of these verses in language X.

To give the reader a flavor of the distinctions that are
reflected in these clusters, we now list phenomena that are
characteristic of verses that contain only one of the five
pivots; these phenomena identify properties of one language
that the other four do not have.

\textbf{CRS ``ti''.} CRS has a set of markers that can be
systematically combined, in particular, 
a progressive marker ``pe'' that can be combined with the
past tense marker ``ti''. As a result, past progressive
sentences in CRS are generally marked with ``ti''.
Example:
``43004031 Meanwhile, the disciples were urging Jesus, `Rabbi, eat
something.''' 
``crs\_bible 43004031 Pandan sa letan, bann disip ti pe sipliy Zezi, `Met!
Manz en pe.'''

The other four languages do not consistently use the pivot
for marking the past progressive; e.g., HWC uses
``was begging'' in 43004031 (instead of ``wen'') and TCS uses ``kip tok
strongwan'' `keep talking strongly' in 43004031 (instead of
``bin'').

\textbf{FIJ ``qai''.} This pivot means ``and then''. It is
highly correlated with past tense in the New Testament
because most sequential descriptions of events are
descriptions of past events. But there are also some
non-past sequences. Example:
``eng\_newliving 44009016 And I will show him how much he must suffer for my name’s
sake.''
``fij\_hindi
44009016
Au na qai vakatakila vua na
levu ni ka e na sota kaya e na vukuqu.''
This verse is future tense, but it continues a temporal
sequence (it starts in the
preceding verse) and therefore FIJ uses ``qai''.
The pivots of the other four languages are not general markers of
temporal sequentiality, so they are not used for the future.

\textbf{HWC ``wen''.} HWC is less explicit than the other
four languages in some respects and more explicit in
others. It is less explicit in that not all sentences in a
sequence of past tense sentences need to be marked
explicitly with ``wen'', resulting in some sentences that
are indistinguishable from present tense. On the other hand,
we found many cases of noun phrases in the other four
languages that refer implicitly to the past, but are
translated as a verb with explicit past tense marking in
HWC. Examples:
``hwc\_2000  
40026046        
Da guy who wen set me up \ldots''
`the guy who WEN set me up', 
``eng\_newliving
40026046 \ldots my betrayer \ldots'';
``hwc\_2000  
43008005 \ldots Moses wen tell us in
da Rules \ldots''
`Moses WEN tell us in the rules', 
``eng\_newliving
43008005 The law of Moses says \ldots'';
``hwc\_2000  47006012 We wen give you guys our
love \ldots'',
``eng\_newliving  47006012 
There is no lack of love on our part \ldots''.
In these cases, the other four languages (and English too)
use a noun phrase with no tense marking that is translated
as a tense-marked clause in HWC.

While preparing this analysis, we realized that HWC ``wen''
unfortunately does not meet one of the criteria we set out for
pivots: it is not unambiguous. In addition to being a past tense
marker (derived from standard English ``went''), it can also
be a conjunction, derived from ``when''. This ambiguity is
the cause for some noise in the clusters marked for presence
of HWC ``wen'' in the figure.

\textbf{TCS ``bin''.} Conditionals is one pattern we found in verses that
are marked with TCS ``bin'', but are not marked for past
tense in the other four languages. Example:
``tcs\_bible 46015046 Wanem i bin kam pas i da
nomal bodi ane den da spiritbodi i bin kam apta.'' `what
came first is the normal body and then the spirit body came after',
``eng\_newliving 46015046 What comes first is
the natural body, then the spiritual body comes later.''
Apparently, ``bin'' also has a modal aspect in TCS: generic
statements that do not refer to specific events are rendered
using ``bin'' in TCS whereas the other four languages (and
also English) use the default unmarked tense, i.e., present
tense.

\textbf{TZO ``laj''.} This pivot indicates perfective
aspect. The other four past tense pivots are not perfective
markers, so that there are verses that are marked with
``laj'', but not marked with the past tense pivots of the
other four languages. Example:
``tzo\_huixtan 40010042 \ldots\ ja'ch-ac'bat bendición yu'un hech laj spas \ldots''
(literally ``a blessing \ldots\ LAJ make''),
``eng\_newliving 40010042 \ldots you will surely be
rewarded.'' Perfective aspect and past are correlated in the
real world since most events that are viewed as simple
wholes are in the past. But  future
events can also be viewed this way as the example shows.

Similar maps for present and future tenses are presented in the \figref{presentmap} and \figref{futuremap}.
\begin{figure*}
\centering
  \includegraphics[width=1\textwidth,trim={8.5cm 6cm 7.5cm 4cm},clip]{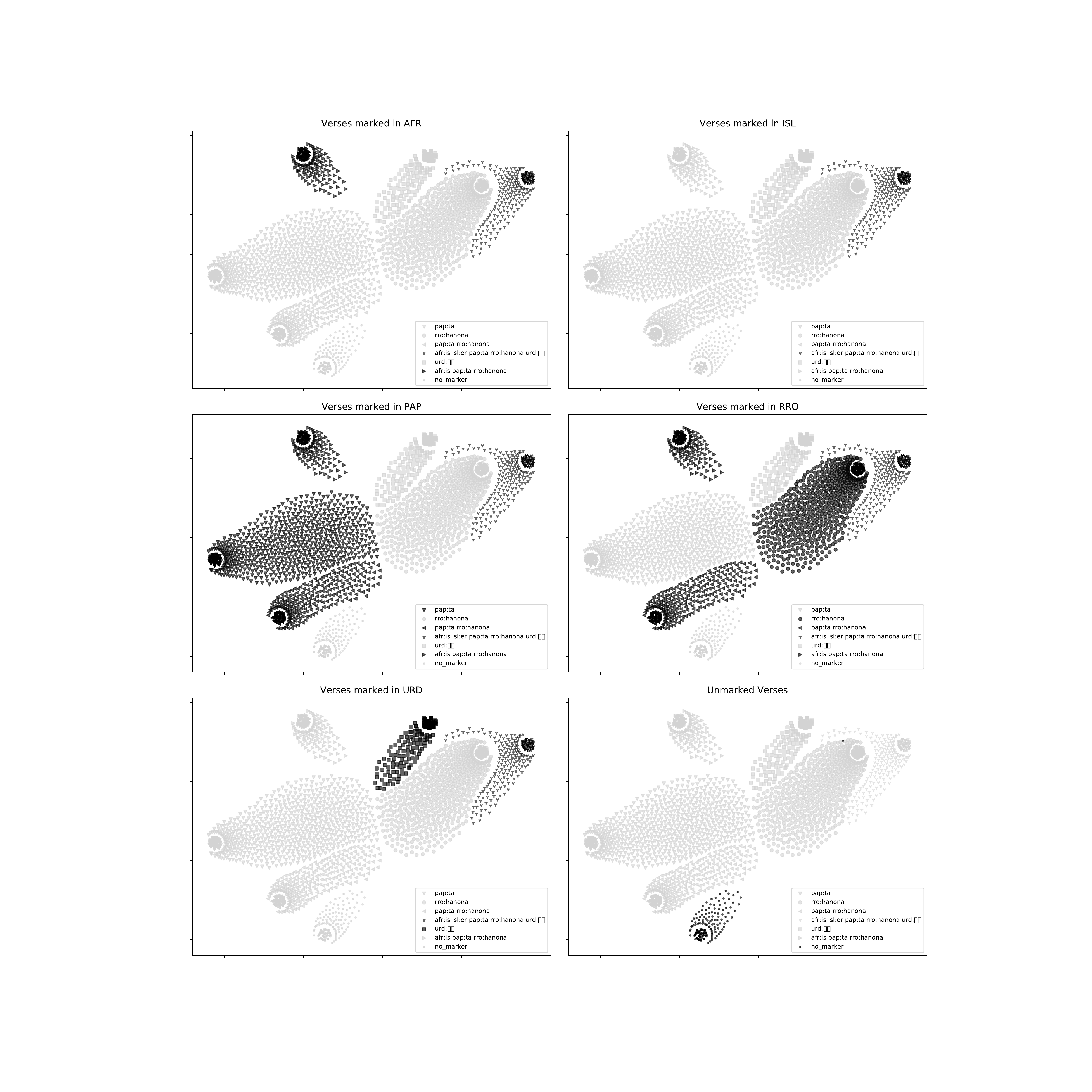}
  \caption{\figlabel{presentmap} A map of present tense based on the
    largest clusters of verses 
with particular combinations of the past tense pivots from
Papiamento (PAP), Waima (RRO), Afrikaans
(ARF), Urdu (URD) and Icelandic (ISL).
For each of the five languages, we present a subfigure
that highlights the subset of verse clusters that are marked
by the pivot of that language. The sixth subfigure
highlights verses not marked by any of the five pivots.}
\end{figure*}

\begin{figure*}
\centering
  \includegraphics[width=1\textwidth,trim={11cm 6cm 10.5cm 5.5cm},clip]{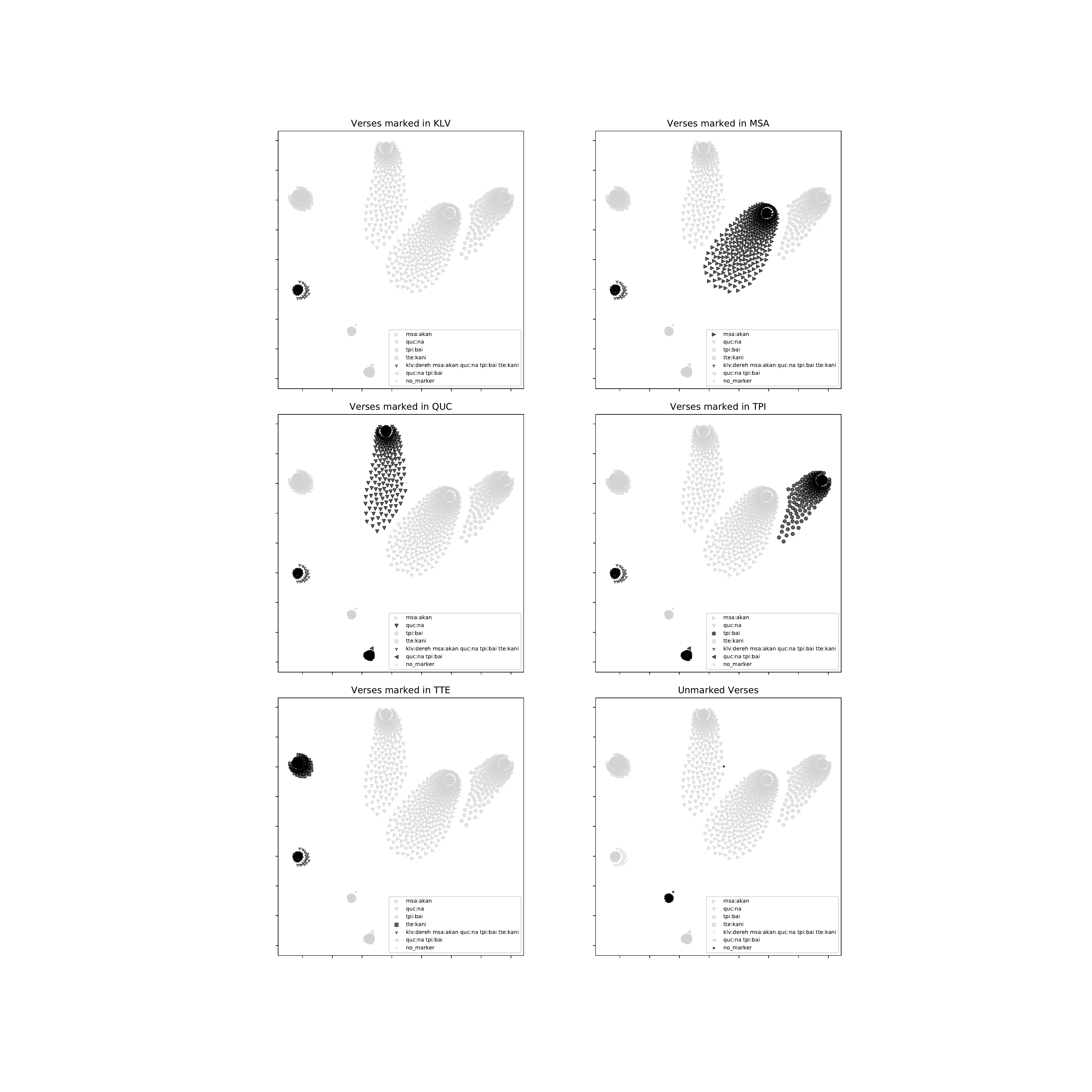}
  \caption{\figlabel{futuremap} A map of future tense based on the
    largest clusters of verses 
with particular combinations of the past tense pivots from
Bwanabwana (TTE), Tok Pisin (TPI), Quiché 
(QUC), Malay (MSA) and Maskelynes (KLV).
For each of the five languages, we present a subfigure
that highlights the subset of verse clusters that are marked
by the pivot of that language. The sixth subfigure
highlights verses not marked by any of the five pivots.}
\end{figure*}

\section{Related work}
\seclabel{related} 
Our work is inspired by
\cite{cysouw2014inducing,cysouw2007parallel}; see also
\cite{dahl2007questionnaires,walchli2010consonant}.
Cysouw creates maps like \figref{pt} by manually identifying
occurrences of the proper noun ``Bible'' in a parallel
corpus of Jehovah's Witnesses' texts. Areas of the map
correspond to semantic roles, e.g., the Bible as actor (it
tells you to do something) or as object (it was
printed). This is a definition of semantic roles that is
complementary to and 
different from prior typological research because it is
empirically grounded in real language use across a large
number of languages. It allows typologists to investigate
traditional questions from a radically new perspective.

The field of \textbf{typology} is
important for both theoretical
\cite{greenberg1960quantitative,whaley1996introduction,croft2002typology}
and computational 
\cite{heiden2000typtex,santaholma2007grammar,bender2009linguistically,bender2011achieving}
linguistics.
Typology is concerned with all areas of linguistics:
morphology \cite{song2014linguistic}, syntax
\cite{comrie1989language,croft2001radical,croft2008inferring,song2014linguistic},
semantic roles
\cite{hartmann2014identifying,cysouw2014inducing}, semantics
\cite{koptjevskaja2007typological,dahl2014perfect,walchli2012lexical,sharma2009typological}
etc.  Typological information is important for many NLP
tasks including discourse analysis
\cite{myhill1992typological}, information retrieval
\cite{pirkola2001morphological}, POS tagging
\cite{bohnet2012transition},  parsing
\cite{bohnet2012transition, mcdonald2013universal}, machine
translation \cite{hajivc2000machine,kunchukuttan2016faster}
and morphology \cite{bohnet2013joint}.

\textbf{Tense} is a central phenomenon in linguistics
and the languages of the world differ greatly in whether and how
they express tense \cite{traugott1978expression,bybee1989creation,dahl2000tense,dahl1985tense,santos2004translation,dahl2007questionnaires,santos2004translation,dahl2014perfect}.

\textbf{Low resource.}  Even resources with the widest
coverage like
World Atlas of Linguistic
Structures (WALS) \cite{dryer2005world} have little
information for hundreds of languages.
Many researchers have
taken advantage of parallel information for
extracting linguistic knowledge in low-resource settings
\cite{resnik1997creating,resnik2004exploiting,mihalcea2005parallel,mayer2014creating,christodouloupoulos2015massively,lison2016opensubtitles2016}.

\subsection{Parallel corpora and annotation projection}
In general, parallel corpora are a resource of immense
importance in natural language processing
at least since
\newcite{brown1993mathematics}'s work
on machine translation and they are
widely used. In addition to machine translation, other
applications include typology
\cite{asgari2016comparing,malaviya17typology} and paraphrase mining
\cite{bannard2005paraphrasing}.

Annotation projection is a specific use of parallel
corpora:
a set of labels that is available for  $L_1$ 
is projected to $L_2$ via alignment links
within the parallel corpus. $L_1$ labels can either be
obtained through manual annotation or through an analysis
module that may be available for $L_1$, but not for $L_2$.
We interpret label here broadly, including, e.g.,  part of speech
labels, morphological tags and segmentation boundaries,
sense labels, mood labels,
event labels, syntactic analysis and coreference.
We can only cite a small subset of papers using annotation
projection published in the last two decades:
\newcite{mcenery1999domains},
\newcite{ide2000cross},
\newcite{yarowsky2001inducing},
\newcite{xiao2002corpus},
\newcite{diab02senseparallel},
\newcite{hwa2005bootstrapping},
\newcite{mukerjee2006detecting},
\newcite{pado09roleprojection},
\newcite{das11posprojection},
\newcite{de2011can},
\newcite{nordrum2015exploring},
\newcite{marasovic16modal} and
\newcite{agic2016multilingual}.

Of particular relevance is work that projects tense:
\newcite{spreyer2008projection},
\newcite{xue13chinesetense},
\newcite{loaiciga2014english},
\newcite{zhang14chinesetense} and
\newcite{friedrich17telicity}.

In contrast to this previous work, the labels we project in
this paper are not the result of human annotation nor the
result of the annotation computed by an NLP analysis
module. Instead we interpret words in $L_1$ as annotation
labels (words like CRS
``ti'' and TZO ``laj'') and project these word annotation labels to
another language $L_2$.

\enote{hs}{this paper:

\cite{chung2016character}

was in the parallel corpus / annotation projectino aprt

i don't undrestand why.

add it back in?}

\section{Discussion}

Our motivation is not to develop a method that can
then be applied to many other corpora. Rather, our
motivation is that many of the more than 1000 languages in
the Parallel Bible Corpus are low-resource and
that providing a method for creating the first richly
annotated corpus (through the projection of annotation we
propose) for many of these languages is a significant
contribution.

The original motivation for our approach is provided by
the work of the typologist Michael Cysouw. He created the
same type of annotation as we, but he produced it manually
whereas we use automatic methods.
But the structure of the annotation and its use
in linguistic analysis is the same as what we provide.

The basic idea of the utility of the final outcome of
SuperPivot is that the 1163 languages all richly annotate
each other. As long as there are a few among the 1163
languages that have a clear marker for linguistic feature $f$,
then this marker can be projected to all other languages to
richly annotate them. For any linguistic feature, there is a
good chance that a few languages clearly mark it. Of
course, this small subset of languages will be different for
every linguistic feature.

Thus, even for extremely resource-poor languages for which at
present no annotated resources exist, SuperPivot will make
available richly annotated corpora that should advance
linguistic research on these languages.

\section{Conclusion}
\seclabel{future}
We presented SuperPivot, an analysis method for low-resource languages
that occur in a superparallel corpus, i.e., in a corpus that
contains an order of magnitude more languages than parallel
corpora currently in use.
We showed that SuperPivot
performs well 
  for the crosslingual analysis of the linguistic phenomenon
  of tense. We produced analysis results for more than 1000
  languages, conducting -- to the best of our knowledge -- the largest crosslingual computational study
  performed to date.
We extended existing methodology for leveraging parallel
corpora for typological analysis by overcoming a limiting
assumption of earlier work. 
We only require that a linguistic feature is overtly marked
in a \emph{few} of thousands of languages as opposed to requiring
that it be marked in \emph{all} languages under investigation.

\section{Future directions}

There are at least two future directions that seem
promising to us.
\begin{itemize}
  \item Creating a common map of tense along
  the lines of \figref{pt}, but unifying the three tenses
  \item Addressing shortcomings of the way we compute
    alignments: (i)
generalizing character n-grams to more general
features, so that templates in templatic morphology,
reduplication and other more complex manifestations of
linguistic features can be captured; (ii)
use n-gram features of different lengths to account for
differences among languages, e.g., 
shorter ones for Chinese, longer ones for
English; (iii)
segmenting
verses into clauses and performing alignment not
on the verse level (which caused
many errors in our experiments),
but on the
clause level instead; (iv) using global information more
effectively, e.g., by 
extracting alignment features from automatically induced bi- or multilingual lexicons.
\end{itemize}

\section*{Acknowledgments}
We gratefully acknowledge
financial support from Volkswagenstiftung and 
fruitful discussions with
Fabienne Braune,
Michael Cysouw,
Alexander Fraser,
Annemarie Friedrich,
Mohsen Mahdavi Mazdeh,
Mohammad R.K. Mofrad,
Yadollah Yaghoobzadeh,
and Benjamin Roth. We are indebted to
Michael Cysouw for 
the Parallel Bible Corpus.


\begin{thebibliography}{}
\expandafter\ifx\csname natexlab\endcsname\relax\def\natexlab#1{#1}\fi

\bibitem[{Agi{\'c} et~al.(2016)Agi{\'c}, Johannsen, Plank, Mart{\'\i}nez,
  Schluter, and S{\o}gaard}]{agic2016multilingual}
{\v{Z}}eljko Agi{\'c}, Anders Johannsen, Barbara Plank, H{\'e}ctor~Alonso
  Mart{\'\i}nez, Natalie Schluter, and Anders S{\o}gaard. 2016.
\newblock Multilingual projection for parsing truly low-resource languages.
\newblock {\em Transactions of the Association for Computational Linguistics\/}
  4:301--312.

\bibitem[{Aissen(1987)}]{aissen1987tzotzil}
Judith~L. Aissen. 1987.
\newblock {\em Tzotzil Clause Structure\/}.
\newblock Springer.

\bibitem[{Andersen(1990)}]{andersen1990papiamentu}
Roger~W Andersen. 1990.
\newblock Papiamentu tense-aspect, with special attention to discourse.
\newblock {\em Pidgin and creole tense-mood-aspect systems\/} pages 59--96.

\bibitem[{Asgari and Mofrad(2016)}]{asgari2016comparing}
Ehsaneddin Asgari and Mohammad R.~K. Mofrad. 2016.
\newblock Comparing fifty natural languages and twelve genetic languages using
  word embedding language divergence {(WELD)} as a quantitative measure of
  language distance.
\newblock In {\em Workshop on Multilingual and Cross-lingual Methods in NLP\/}.
  pages 65--74.

\bibitem[{Bannard and Callison-Burch(2005)}]{bannard2005paraphrasing}
Colin Bannard and Chris Callison-Burch. 2005.
\newblock Paraphrasing with bilingual parallel corpora.
\newblock In {\em Proceedings of the 43rd Annual Meeting on Association for
  Computational Linguistics\/}. Association for Computational Linguistics,
  pages 597--604.

\bibitem[{Bender(2009)}]{bender2009linguistically}
Emily~M Bender. 2009.
\newblock Linguistically na{\"\i}ve!= language independent: why nlp needs
  linguistic typology.
\newblock In {\em Proceedings of the EACL 2009 Workshop on the Interaction
  between Linguistics and Computational Linguistics: Virtuous, Vicious or
  Vacuous?\/}. Association for Computational Linguistics, pages 26--32.

\bibitem[{Bender(2011)}]{bender2011achieving}
Emily~M Bender. 2011.
\newblock On achieving and evaluating language-independence in nlp.
\newblock {\em Linguistic Issues in Language Technology\/} 6(3):1--26.

\bibitem[{Bird(2006)}]{bird2006nltk}
Steven Bird. 2006.
\newblock Nltk: the natural language toolkit.
\newblock In {\em Proceedings of the COLING/ACL on Interactive presentation
  sessions\/}. Association for Computational Linguistics, pages 69--72.

\bibitem[{Bohnet and Nivre(2012)}]{bohnet2012transition}
Bernd Bohnet and Joakim Nivre. 2012.
\newblock A transition-based system for joint part-of-speech tagging and
  labeled non-projective dependency parsing.
\newblock In {\em Proceedings of the 2012 Joint Conference on Empirical Methods
  in Natural Language Processing and Computational Natural Language
  Learning\/}. Association for Computational Linguistics, pages 1455--1465.

\bibitem[{Bohnet et~al.(2013)Bohnet, Nivre, Boguslavsky, Farkas, Ginter, and
  Haji{\v{c}}}]{bohnet2013joint}
Bernd Bohnet, Joakim Nivre, Igor Boguslavsky, Rich{\'a}rd Farkas, Filip Ginter,
  and Jan Haji{\v{c}}. 2013.
\newblock Joint morphological and syntactic analysis for richly inflected
  languages.
\newblock {\em Transactions of the Association for Computational Linguistics\/}
  1:415--428.

\bibitem[{Brown et~al.(1993)Brown, Pietra, Pietra, and
  Mercer}]{brown1993mathematics}
Peter~F Brown, Vincent J~Della Pietra, Stephen A~Della Pietra, and Robert~L
  Mercer. 1993.
\newblock The mathematics of statistical machine translation: Parameter
  estimation.
\newblock {\em Computational linguistics\/} 19(2):263--311.

\bibitem[{Bybee and Dahl(1989)}]{bybee1989creation}
Joan~L Bybee and {\"O}sten Dahl. 1989.
\newblock {\em The creation of tense and aspect systems in the languages of the
  world\/}.
\newblock John Benjamins Amsterdam.

\bibitem[{Casella and Berger(2008)}]{casella08statistical}
George Casella and Roger~L. Berger. 2008.
\newblock {\em Statistical Inference\/}.
\newblock Thomson.

\bibitem[{Christodouloupoulos and
  Steedman(2015)}]{christodouloupoulos2015massively}
Christos Christodouloupoulos and Mark Steedman. 2015.
\newblock A massively parallel corpus: the bible in 100 languages.
\newblock {\em Language resources and evaluation\/} 49(2):375--395.

\bibitem[{Comrie(1989)}]{comrie1989language}
Bernard Comrie. 1989.
\newblock {\em Language universals and linguistic typology: Syntax and
  morphology\/}.
\newblock University of Chicago press.

\bibitem[{Croft(2001)}]{croft2001radical}
William Croft. 2001.
\newblock {\em Radical construction grammar: Syntactic theory in typological
  perspective\/}.
\newblock Oxford University Press on Demand.

\bibitem[{Croft(2002)}]{croft2002typology}
William Croft. 2002.
\newblock {\em Typology and universals\/}.
\newblock Cambridge University Press.

\bibitem[{Croft and Poole(2008)}]{croft2008inferring}
William Croft and Keith~T Poole. 2008.
\newblock Inferring universals from grammatical variation: Multidimensional
  scaling for typological analysis.
\newblock {\em Theoretical linguistics\/} 34(1):1--37.

\bibitem[{Cysouw(2014)}]{cysouw2014inducing}
Michael Cysouw. 2014.
\newblock Inducing semantic roles.
\newblock {\em Perspectives on semantic roles\/} pages 23--68.

\bibitem[{Cysouw and W{\"a}lchli(2007)}]{cysouw2007parallel}
Michael Cysouw and Bernhard W{\"a}lchli. 2007.
\newblock Parallel texts: using translational equivalents in linguistic
  typology.
\newblock {\em STUF-Sprachtypologie und Universalienforschung\/} 60(2):95--99.

\bibitem[{Dahl(1985)}]{dahl1985tense}
{\"O}sten Dahl. 1985.
\newblock {\em Tense and aspect systems\/}.
\newblock Basil Blackwell.

\bibitem[{Dahl(2000)}]{dahl2000tense}
{\"O}sten Dahl. 2000.
\newblock {\em Tense and Aspect in the Languages of Europe\/}.
\newblock Walter de Gruyter.

\bibitem[{Dahl(2007)}]{dahl2007questionnaires}
{\"O}sten Dahl. 2007.
\newblock From questionnaires to parallel corpora in typology.
\newblock {\em STUF-Sprachtypologie und Universalienforschung\/}
  60(2):172--181.

\bibitem[{Dahl(2014)}]{dahl2014perfect}
{\"O}sten Dahl. 2014.
\newblock The perfect map: Investigating the cross-linguistic distribution of
  tame categories in a parallel corpus.
\newblock {\em Aggregating Dialectology, Typology, and Register Contents
  Analysis. Linguistic Variation in Text and Speech. Linguae \& litterae\/}
  28:268--289.

\bibitem[{Das and Petrov(2011)}]{das11posprojection}
Dipanjan Das and Slav Petrov. 2011.
\newblock Unsupervised part-of-speech tagging with bilingual graph-based
  projections.
\newblock In {\em The 49th Annual Meeting of the Association for Computational
  Linguistics: Human Language Technologies, Proceedings of the Conference,
  19-24 June, 2011, Portland, Oregon, {USA}\/}. pages 600--609.

\bibitem[{de~Souza and Or{\u{a}}san(2011)}]{de2011can}
Jos{\'e} Guilherme~Camargo de~Souza and Constantin Or{\u{a}}san. 2011.
\newblock Can projected chains in parallel corpora help coreference resolution?
\newblock In {\em Discourse Anaphora and Anaphor Resolution Colloquium\/}.
  Springer, pages 59--69.

\bibitem[{Diab and Resnik(2002)}]{diab02senseparallel}
Mona~T. Diab and Philip Resnik. 2002.
\newblock An unsupervised method for word sense tagging using parallel corpora.
\newblock In {\em Proceedings of the 40th Annual Meeting of the Association for
  Computational Linguistics, July 6-12, 2002, Philadelphia, PA, {USA.}\/}.
  pages 255--262.

\bibitem[{Dryer et~al.(2005)Dryer, Gil, Comrie, Jung, Schmidt
  et~al.}]{dryer2005world}
Matthew~S Dryer, David Gil, Bernard Comrie, Hagen Jung, Claudia Schmidt, et~al.
  2005.
\newblock {\em The world atlas of language structures\/}.
\newblock Oxford University Press.

\bibitem[{Dyer et~al.(2013)Dyer, Chahuneau, and Smith}]{dyer2013simple}
Chris Dyer, Victor Chahuneau, and Noah~A. Smith. 2013.
\newblock A simple, fast, and effective reparameterization of {IBM} model 2.
\newblock In {\em Human Language Technologies: Conference of the North American
  Chapter of the Association of Computational Linguistics, Proceedings, June
  9-14, 2013, Westin Peachtree Plaza Hotel, Atlanta, Georgia, {USA}\/}. pages
  644--648.

\bibitem[{Friedrich and Gateva(2017)}]{friedrich17telicity}
Annemarie Friedrich and Damyana Gateva. 2017.
\newblock Classification of telicity using cross-linguistic annotation
  projection.
\newblock In {\em Proceedings of the 2017 Conference on Empirical Methods in
  Natural Language Processing\/}. Association for Computational Linguistics,
  Copenhagen, Denmark, pages 2549--2555.

\bibitem[{Greenberg(1960)}]{greenberg1960quantitative}
Joseph~H Greenberg. 1960.
\newblock A quantitative approach to the morphological typology of language.
\newblock {\em International journal of American linguistics\/} 26(3):178--194.

\bibitem[{Haji{\v{c}} et~al.(2000)Haji{\v{c}}, Hric, and
  Kubo{\v{n}}}]{hajivc2000machine}
Jan Haji{\v{c}}, Jan Hric, and Vladislav Kubo{\v{n}}. 2000.
\newblock Machine translation of very close languages.
\newblock In {\em Proceedings of the sixth conference on Applied natural
  language processing\/}. Association for Computational Linguistics, pages
  7--12.

\bibitem[{Hartmann et~al.(2014)Hartmann, Haspelmath, and
  Cysouw}]{hartmann2014identifying}
Iren Hartmann, Martin Haspelmath, and Michael Cysouw. 2014.
\newblock Identifying semantic role clusters and alignment types via microrole
  coexpression tendencies.
\newblock {\em Studies in Language. International Journal sponsored by the
  Foundation ?Foundations of Language?\/} 38(3):463--484.

\bibitem[{Heiden et~al.(2000)Heiden, Pr{\'e}vost, Habert, Folch, Fleury,
  Illouz, Lafon, and Nioche}]{heiden2000typtex}
Serge Heiden, Sophie Pr{\'e}vost, Benoit Habert, Helka Folch, Serge Fleury,
  Gabriel Illouz, Pierre Lafon, and Julien Nioche. 2000.
\newblock Typtex: Inductive typological text classification by multivariate
  statistical analysis for nlp systems tuning/evaluation.
\newblock In {\em Maria Gavrilidou, George Carayannis, Stella Markantonatou,
  Stelios Piperidis, Gregory Stainhaouer ({\'e}ds) Second International
  Conference on Language Resources and Evaluation\/}. pages p--141.

\bibitem[{Hwa et~al.(2005)Hwa, Resnik, Weinberg, Cabezas, and
  Kolak}]{hwa2005bootstrapping}
Rebecca Hwa, Philip Resnik, Amy Weinberg, Clara Cabezas, and Okan Kolak. 2005.
\newblock Bootstrapping parsers via syntactic projection across parallel texts.
\newblock {\em Natural language engineering\/} 11(03):311--325.

\bibitem[{Ide(2000)}]{ide2000cross}
Nancy Ide. 2000.
\newblock Cross-lingual sense determination: Can it work?
\newblock {\em Computers and the Humanities\/} 34(1):223--234.

\bibitem[{Johnson(1967)}]{johnson1967hierarchical}
Stephen~C Johnson. 1967.
\newblock Hierarchical clustering schemes.
\newblock {\em Psychometrika\/} 32(3):241--254.

\bibitem[{Koptjevskaja-Tamm et~al.(2007)Koptjevskaja-Tamm, Vanhove, and
  Koch}]{koptjevskaja2007typological}
Maria Koptjevskaja-Tamm, Martine Vanhove, and Peter Koch. 2007.
\newblock Typological approaches to lexical semantics.
\newblock {\em Linguistic typology\/} 11(1):159--185.

\bibitem[{Kunchukuttan and Bhattacharyya(2016)}]{kunchukuttan2016faster}
Anoop Kunchukuttan and Pushpak Bhattacharyya. 2016.
\newblock Faster decoding for subword level phrase-based smt between related
  languages.
\newblock {\em arXiv preprint arXiv:1611.00354\/} .

\bibitem[{Lison and Tiedemann(2016)}]{lison2016opensubtitles2016}
Pierre Lison and J{\"o}rg Tiedemann. 2016.
\newblock Opensubtitles2016: Extracting large parallel corpora from movie and
  tv subtitles.
\newblock In {\em Proceedings of the 10th International Conference on Language
  Resources and Evaluation\/}.

\bibitem[{Loaiciga et~al.(2014)Loaiciga, Meyer, and
  Popescu-Belis}]{loaiciga2014english}
Sharid Loaiciga, Thomas Meyer, and Andrei Popescu-Belis. 2014.
\newblock English-french verb phrase alignment in europarl for tense
  translation modeling.
\newblock In {\em The Ninth Language Resources and Evaluation Conference\/}.
  EPFL-CONF-198442.

\bibitem[{Maaten and Hinton(2008)}]{maaten2008visualizing}
Laurens van~der Maaten and Geoffrey Hinton. 2008.
\newblock Visualizing data using t-sne.
\newblock {\em Journal of Machine Learning Research\/} 9(Nov):2579--2605.

\bibitem[{Malaviya et~al.(2017)Malaviya, Neubig, and
  Littell}]{malaviya17typology}
Chaitanya Malaviya, Graham Neubig, and Patrick Littell. 2017.
\newblock Learning language representations for typology prediction.
\newblock In {\em Proceedings of the 2017 Conference on Empirical Methods in
  Natural Language Processing\/}. Association for Computational Linguistics,
  Copenhagen, Denmark, pages 2519--2525.

\bibitem[{Marasovi\'{c} and Frank(2016)}]{marasovic16modal}
Ana Marasovi\'{c} and Anette Frank. 2016.
\newblock Multilingual modal sense classification using a convolutional neural
  network.
\newblock In {\em Workshop on Representation Learning for NLP\/}. pages
  111--120.

\bibitem[{Mayer and Cysouw(2014)}]{mayer2014creating}
Thomas Mayer and Michael Cysouw. 2014.
\newblock Creating a massively parallel bible corpus.
\newblock {\em Oceania\/} 135(273):40.

\bibitem[{McDonald et~al.(2013)McDonald, Nivre, Quirmbach-Brundage, Goldberg,
  Das, Ganchev, Hall, Petrov, Zhang, T{\"a}ckstr{\"o}m
  et~al.}]{mcdonald2013universal}
Ryan~T McDonald, Joakim Nivre, Yvonne Quirmbach-Brundage, Yoav Goldberg,
  Dipanjan Das, Kuzman Ganchev, Keith~B Hall, Slav Petrov, Hao Zhang, Oscar
  T{\"a}ckstr{\"o}m, et~al. 2013.
\newblock Universal dependency annotation for multilingual parsing.
\newblock In {\em ACL (2)\/}. pages 92--97.

\bibitem[{McEnery and Xiao(1999)}]{mcenery1999domains}
Tony McEnery and Richard Xiao. 1999.
\newblock Domains, text types, aspect marking and english-chinese translation.
\newblock {\em Languages in Contrast\/} 2(2):211--229.

\bibitem[{McWhorter(2005)}]{mcwhorter2005defining}
John~H McWhorter. 2005.
\newblock {\em Defining creole\/}.
\newblock Oxford University Press.

\bibitem[{Mihalcea and Simard(2005)}]{mihalcea2005parallel}
Rada Mihalcea and Michel Simard. 2005.
\newblock Parallel texts.
\newblock {\em Natural Language Engineering\/} 11(03):239--246.

\bibitem[{Mukerjee et~al.(2006)Mukerjee, Soni, and
  Raina}]{mukerjee2006detecting}
Amitabha Mukerjee, Ankit Soni, and Achla~M Raina. 2006.
\newblock Detecting complex predicates in hindi using pos projection across
  parallel corpora.
\newblock In {\em Proceedings of the Workshop on Multiword Expressions:
  Identifying and Exploiting Underlying Properties\/}. Association for
  Computational Linguistics, pages 28--35.

\bibitem[{Myhill and Myhill(1992)}]{myhill1992typological}
John Myhill and Myhill. 1992.
\newblock {\em Typological discourse analysis: Quantitative approaches to the
  study of linguistic function\/}.
\newblock Blackwell Oxford.

\bibitem[{Nordrum(2015)}]{nordrum2015exploring}
Lene Nordrum. 2015.
\newblock Exploring spontaneous-event marking though parallel corpora:
  Translating english ergative intransitive constructions into norwegian and
  swedish.
\newblock {\em Languages in Contrast\/} 15(2):230--250.

\bibitem[{Pad{\'{o}} and Lapata(2009)}]{pado09roleprojection}
Sebastian Pad{\'{o}} and Mirella Lapata. 2009.
\newblock \href{https://doi.org/10.1613/jair.2863}{Cross-lingual annotation
  projection for semantic roles}.
\newblock {\em J. Artif. Intell. Res.\/} 36:307--340.
\newblock
  \href{https://doi.org/10.1613/jair.2863}{https://doi.org/10.1613/jair.2863}.

\bibitem[{Pirkola(2001)}]{pirkola2001morphological}
Ari Pirkola. 2001.
\newblock Morphological typology of languages for ir.
\newblock {\em Journal of Documentation\/} 57(3):330--348.

\bibitem[{Resnik(2004)}]{resnik2004exploiting}
Philip Resnik. 2004.
\newblock Exploiting hidden meanings: Using bilingual text for monolingual
  annotation.
\newblock {\em Computational Linguistics and Intelligent Text Processing\/}
  pages 283--299.

\bibitem[{Resnik et~al.(1997)Resnik, Olsen, and Diab}]{resnik1997creating}
Philip Resnik, Mari~Broman Olsen, and Mona Diab. 1997.
\newblock Creating a parallel corpus from the book of 2000 tongues.
\newblock In {\em Proceedings of the Text Encoding Initiative 10th Anniversary
  User Conference (TEI-10)\/}. Citeseer.

\bibitem[{Sankoff(1990)}]{sankoff1990grammaticalization}
Gillian Sankoff. 1990.
\newblock The grammaticalization of tense and aspect in tok pisin and sranan.
\newblock {\em Language Variation and Change\/} 2(03):295--312.

\bibitem[{Santaholma(2007)}]{santaholma2007grammar}
Marianne~Elina Santaholma. 2007.
\newblock Grammar sharing techniques for rule-based multilingual nlp systems.
\newblock {\em Proceedings of the 16th Nordic Conference of Computational
  Linguistics (NODALIDA)\/} .

\bibitem[{Santos(2004)}]{santos2004translation}
Diana Santos. 2004.
\newblock {\em Translation-based corpus studies: Contrasting English and
  Portuguese tense and aspect systems\/}.
\newblock 50. Rodopi.

\bibitem[{Sharma(2009)}]{sharma2009typological}
Devyani Sharma. 2009.
\newblock Typological diversity in new englishes.
\newblock {\em English World-Wide\/} 30(2):170--195.

\bibitem[{Song(2014)}]{song2014linguistic}
Jae~Jung Song. 2014.
\newblock {\em Linguistic typology: Morphology and syntax\/}.
\newblock Routledge.

\bibitem[{Spreyer and Frank(2008)}]{spreyer2008projection}
Kathrin Spreyer and Anette Frank. 2008.
\newblock Projection-based acquisition of a temporal labeller.
\newblock In {\em IJCNLP\/}. pages 489--496.

\bibitem[{Traugott(1978)}]{traugott1978expression}
Elizabeth~Closs Traugott. 1978.
\newblock On the expression of spatio-temporal relations in language.
\newblock {\em Universals of human language\/} 3:369--400.

\bibitem[{W{\"a}lchli(2010)}]{walchli2010consonant}
Bernhard W{\"a}lchli. 2010.
\newblock The consonant template in synchrony and diachrony.
\newblock {\em Baltic linguistics\/} 1.

\bibitem[{W{\"a}lchli and Cysouw(2012)}]{walchli2012lexical}
Bernhard W{\"a}lchli and Michael Cysouw. 2012.
\newblock Lexical typology through similarity semantics: Toward a semantic map
  of motion verbs.
\newblock {\em Linguistics\/} 50(3):671--710.

\bibitem[{Whaley(1996)}]{whaley1996introduction}
Lindsay~J Whaley. 1996.
\newblock {\em Introduction to typology: the unity and diversity of
  language\/}.
\newblock Sage Publications.

\bibitem[{Xiao and McEnery(2002)}]{xiao2002corpus}
RZ~Xiao and AM~McEnery. 2002.
\newblock A corpus-based approach to tense and aspect in english-chinese
  translation.
\newblock In {\em The 1st International Symposium on Contrastive and
  Translation Studies between Chinese and English\/}.

\bibitem[{Xue et~al.(2013)Xue, Zhang, and Yang}]{xue13chinesetense}
Nianwen Xue, Yuchen Zhang, and Yaqin Yang. 2013.
\newblock
  \href{http://aclanthology.coli.uni-saarland.de/pdf/W/W13/W13-0307.pdf}{Distant
  annotation of chinese tense and modality}.
\newblock In {\em Workshop on Annotation of Modal Meaning in Natural Language
  (WAMM)\/}. Association for Computational Linguistics, pages 47--55.
\newblock
  \href{http://aclanthology.coli.uni-saarland.de/pdf/W/W13/W13-0307.pdf}{http://aclanthology.coli.uni-saarland.de/pdf/W/W13/W13-0307.pdf}.

\bibitem[{Yarowsky et~al.(2001)Yarowsky, Ngai, and
  Wicentowski}]{yarowsky2001inducing}
David Yarowsky, Grace Ngai, and Richard Wicentowski. 2001.
\newblock Inducing multilingual text analysis tools via robust projection
  across aligned corpora.
\newblock In {\em Proceedings of the first international conference on Human
  language technology research\/}. Association for Computational Linguistics,
  pages 1--8.

\bibitem[{Zhang and Xue(2014)}]{zhang14chinesetense}
Yuchen Zhang and Nianwen Xue. 2014.
\newblock \href{http://www.aclweb.org/anthology/D14-1204}{Automatic inference
  of the tense of chinese events using implicit linguistic information}.
\newblock In {\em Proceedings of the 2014 Conference on Empirical Methods in
  Natural Language Processing (EMNLP)\/}. Association for Computational
  Linguistics, Doha, Qatar, pages 1902--1911.
\newblock
  \href{http://www.aclweb.org/anthology/D14-1204}{http://www.aclweb.org/anthology/D14-1204}.

\end{thebibliography}
\end{document}